\documentclass[sigconf,article]{acmart}

\renewcommand\footnotetextcopyrightpermission[1]{} 

\usepackage{float}
\usepackage{booktabs}
\usepackage{subcaption}
\usepackage{diagbox}
\usepackage{xcolor}
\usepackage{pifont} 
\usepackage{enumitem}
\usepackage{makecell}
\usepackage[most]{tcolorbox} 

\newcommand{\numedit}[1]{\textcolor{black}{#1}}

\begin{document}

\title{Overview of the NTCIR-19 SciClaimEval Task:
Cross-modal Claim Verification in Scientific Papers}

\author{Xanh Ho*}
\affiliation{
    \institution{NII LLMC, Japan}
     \country{}
}
\email{xanh@nii.ac.jp}
\author{Andr\'{e} Greiner-Petter*}
\affiliation{
    \institution{University of Göttingen, Germany}
    \country{}
}
\email{greinerpetter@gipplab.org}
\author{Florian Boudin}
\affiliation{
    \institution{Inria, LS2N, Nantes Université, France}
    \country{}
}
\email{florian.boudin@univ-nantes.fr}
\author{Yun-Ang Wu}
\affiliation{
    \institution{NII LLMC, Japan}
    \country{}
}
\email{yunangwu@nii.ac.jp}
\author{Sunisth Kumar}
\affiliation{
    \institution{The University of Tokyo, Japan}
    \country{}
}
\email{sunisth@g.ecc.u-tokyo.ac.jp}
\author{Tian Cheng Xia}
\affiliation{
    \institution{University of Bologna, Italy}
    \country{}
}
\email{tiancheng.xia@studio.unibo.it}
\author{Hiba Akbi}
\affiliation{
    \institution{Sorbonne University, France}
    \country{}
}
\email{Hiba.Akbi@etu.sorbonne-universite.fr}
\author{Tim Schopf}
\affiliation{
    \institution{NII, Japan}
    \country{}
}
\email{tim.schopf@t-online.de}

\author{Akiko Aizawa}
\affiliation{
    \institution{NII, Japan}
    \country{}
}
\email{aizawa@nii.ac.jp}

\begin{abstract}
This paper presents an overview of \emph{SciClaimEval}, a pilot task at NTCIR-19 on verifying scientific claims against the tables and figures of the papers in which they appear.
Our dataset pairs authentic claims from 180 papers in machine learning, natural language processing, and medicine with contrastive versions of their tables and figures, one supporting the claim and one minimally modified so that it does not.
The task comprises two subtasks: deciding whether a piece of evidence supports a claim, and selecting which of two evidence versions does.
Twenty-two teams registered, the highest number among NTCIR-19 tasks, and 14 submitted 177 runs.
Most systems built on multimodal large language models, relying on prompting and multi-stage inference rather than task-specific fine-tuning.
On the first subtask, the best systems reached 98.1\% pair accuracy, but all six top-ranked systems relied on the contrastive pairs; the best system that judged each claim-evidence pair on its own scored 24.3 points lower.
Open-weight models, embedded in carefully designed pipelines, matched proprietary ones.
Yet verifying a single claim against its evidence, the setting that matters in practice, remains an open problem.
%
%
%
%

\end{abstract}

\keywords{Claim Verification, Cross-modal, Scientific Papers}

\maketitle
\pagestyle{plain} 




\section{Introduction}

\begingroup\def\thefootnote{*}\footnotetext{Equal contribution.}\endgroup

Scientific papers make claims that should be supported by the evidence they present, typically results reported in tables and figures.
\emph{Scientific claim verification} asks whether a claim is adequately supported by this evidence~\cite{wadden-etal-2020-fact,wang-etal-2021-semeval,lu-etal-2023-scitab}. 
Unlike general fact-checking~\cite{thorne-etal-2018-fever,jiang-etal-2020-hover}, which verifies claims against external knowledge, it focuses on the internal consistency between a paper's claims and its own evidence.
%
%
Checking this consistency is a fundamental part of peer review, yet it is time-consuming and increasingly difficult as submission volumes grow.
Automatic verification could assist reviewers, help authors find unsupported claims before submission, and ultimately improve the reliability of published research.

Existing datasets for scientific claim verification differ mainly in the evidence they consider.
%
%
Tables are the most common, either in structured formats such as JSON or XML, as in SciTab~\cite{lu-etal-2023-scitab}, SEM-TAB-FACTs~\cite{wang-etal-2021-semeval}, and SciAtomicBench~\cite{zhang2025atomicreasoningscientifictable}, or as images, as in MMSci-Eval~\cite{yang2025doestablesourcematter}.
Others turn to figures, such as MuSciClaims~\cite{lal-etal-2025-musciclaims}, whose figures often contain complex subplots and embedded tables, while SciVer~\cite{wang-etal-2025-sciver} considers multiple charts, tables, and text paragraphs as evidence.
%
%
Despite this progress, three limitations remain.
Refuted examples are usually created by editing the claim, for instance by flipping its meaning, which can leave surface cues that models exploit instead of reasoning over the evidence.
Many datasets also rely on claims that are not taken directly from scientific papers.
Finally, tables are typically available in a single format, so it remains unclear how well models handle the same content across different representations.

\begin{figure*}[h]
\centering
    \includegraphics[width=0.93\textwidth,trim={6.7cm 6.8cm 8.3cm 4.8cm},clip]{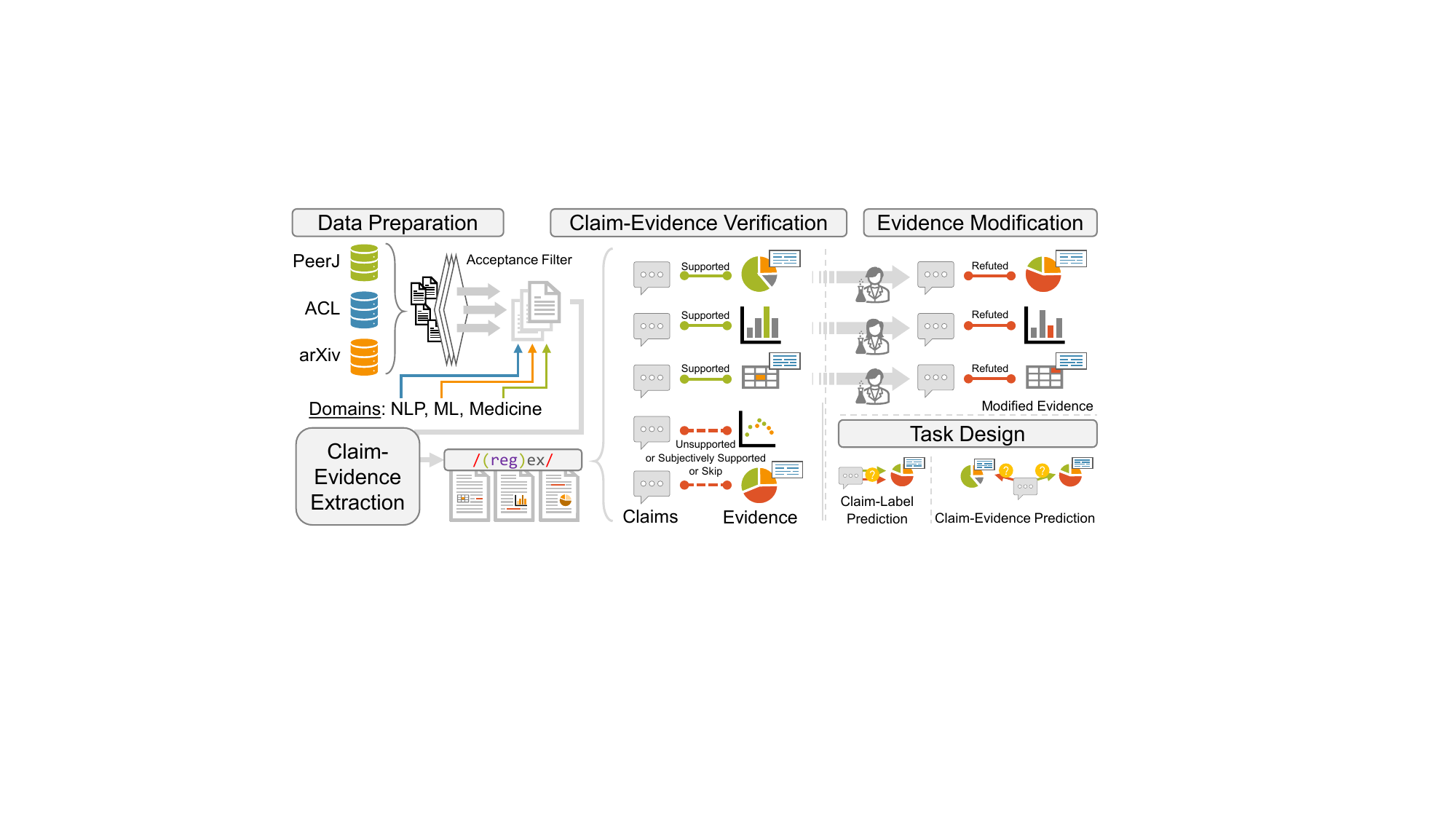}
    \caption{
    Our dataset construction pipeline consists of three main steps: data preparation, automatic claim-evidence extraction, and human annotation. The human annotation process involves two tasks: claim-evidence verification and evidence modification.
    After collecting all samples, we design two subtasks in our dataset: claim-label prediction and claim-evidence prediction. 
    }
    \label{fig_overall}
\end{figure*}

With these limitations in mind, we organized \emph{SciClaimEval}, a pilot task at NTCIR-19 on verifying scientific claims against the tables and figures of the papers in which they appear.
It complements earlier shared tasks on scientific claim verification, which used tables (SemEval-2021 Task 9~\cite{wang-etal-2021-semeval}), abstracts (SCIVER~\cite{wadden-lo-2021-overview}), or social media posts linked to their source papers (CheckThat!~\cite{schellhammer2026overview}).
Rather than editing claims, we create refuted examples by having annotators modify the evidence.
Each claim is thus paired with two versions of its evidence, one that supports it and one that does not, so the wording of the claim carries no cue about the label.
All claims are authentic sentences taken from published papers, and tables are released in four formats: images, LaTeX, HTML, and JSON.

The SciClaimEval dataset is built in three steps.
We first collect peer-reviewed papers in natural language processing (NLP) and machine learning (ML), accepted at ACL venues and NeurIPS, and in medicine, published in PeerJ.
%
From each paper, we extract the text, tables, and figures.
We then identify sentences that explicitly reference a table or figure (e.g., ``Table 1''), as well as other sentences in the same paragraph, and pair each sentence individually with the referenced evidence.
Finally, annotators check whether each claim is supported by its evidence and, for supported claims, edit the evidence to create an unsupported counterpart.
%
%
%
The resulting dataset contains 1,633 claims from 180 papers, with both table- and figure-based evidence across all three domains (Section~\ref{subsec:dataset-creation}).

The pilot task comprises two subtasks.
In Subtask 1, systems decide whether a given piece of evidence supports a claim.
In Subtask 2, systems receive a claim with its two evidence versions and must select the one that supports it.
Twenty-two teams registered, the highest number among NTCIR-19 tasks, and 14 of them submitted 177 runs in total: 106 for Subtask 1 and 71 for Subtask 2.

Most systems built on multimodal large language models (MLLMs).
For Subtask 1, they followed two broad strategies. 
Some verified each claim-evidence pair independently, often with claim decomposition, structured extraction of evidence from images, or executable checks.
Others exploited the paired structure of the data by comparing the two evidence versions of the same claim.
For Subtask 2, systems primarily adopted direct candidate comparison, independent verification followed by selection, or hybrid approaches combining both strategies. Common techniques included ensembles, multi-agent debate, evidence extraction, and consistency checks under reversed candidate order.

We draw the following lessons from this pilot task:
\begin{itemize}
    \item \textbf{High scores rely on paired evidence.} 
    On Subtask 1, the best systems reached 98.1\% pair accuracy on the test set by comparing the two evidence versions of each claim.
    Teams that also evaluated their systems without this information observed substantial performance drops.
    On the development set, for instance, removing paired information cost the two top systems 12 and 21 points of pair accuracy.

    \item \textbf{Open-weight models can match proprietary ones.}
    A structured pipeline built around an open-weight model, with selective human review, matched direct prompting of a proprietary model on Subtask 1 (98.1\% pair accuracy for both) and came close on Subtask 2 (98.1\% versus 98.4\% accuracy).

    \item \textbf{Verifying a single claim against its evidence remains an open problem.} In practice, reviewers and authors face one claim and one piece of evidence, with no contrasting version to compare against. This is precisely the setting in which performance drops, and one that our paired design did not measure directly.

\end{itemize}

\section{Task Design and Dataset}

\subsection{Dataset Creation}
\label{subsec:dataset-creation}

Figure~\ref{fig_overall} illustrates our dataset construction pipeline, which consists of three steps: data preparation, automatic claim-evidence extraction, and human annotation.

\paragraph{Data Preparation.}
We collect papers from three domains. 
For NLP, we take papers from the ACL Anthology and map them to their arXiv versions when available.
For ML, we select arXiv papers whose comments indicate acceptance at conferences such as NeurIPS.
For both domains, we retrieve the HTML-rendered papers from ar5iv~\cite{SML:ar5iv:04:2024} and parse them into JSON containing the title, abstract, and section paragraphs.
Figures are extracted from ar5iv as PNG files, and tables are extracted from the LaTeX source and kept in LaTeX format.
%
%
For medicine, we collect papers published in PeerJ between 2024 and 2025 in the Medicine Articles category. We parse their HTML versions into JSON, storing figures as PNG files and tables in HTML format.

\paragraph{Claim-Evidence Extraction.}
We identify paragraphs that refer to a table or figure using regular expressions that match references such as ``Table 1'', ``Tab. 1'', or ``Fig. 2''.
Each such paragraph is split into sentences, and each sentence is paired with the referenced table or figure, yielding a set of candidate claim-evidence pairs.
Not every sentence is a verifiable claim, so these candidates are filtered during human annotation. We consider only the main text, excluding appendices and their tables and figures.


\begin{figure*}[h]
\centering
    \includegraphics[width=0.93\textwidth,trim={0px 2px 4px 0px},clip]{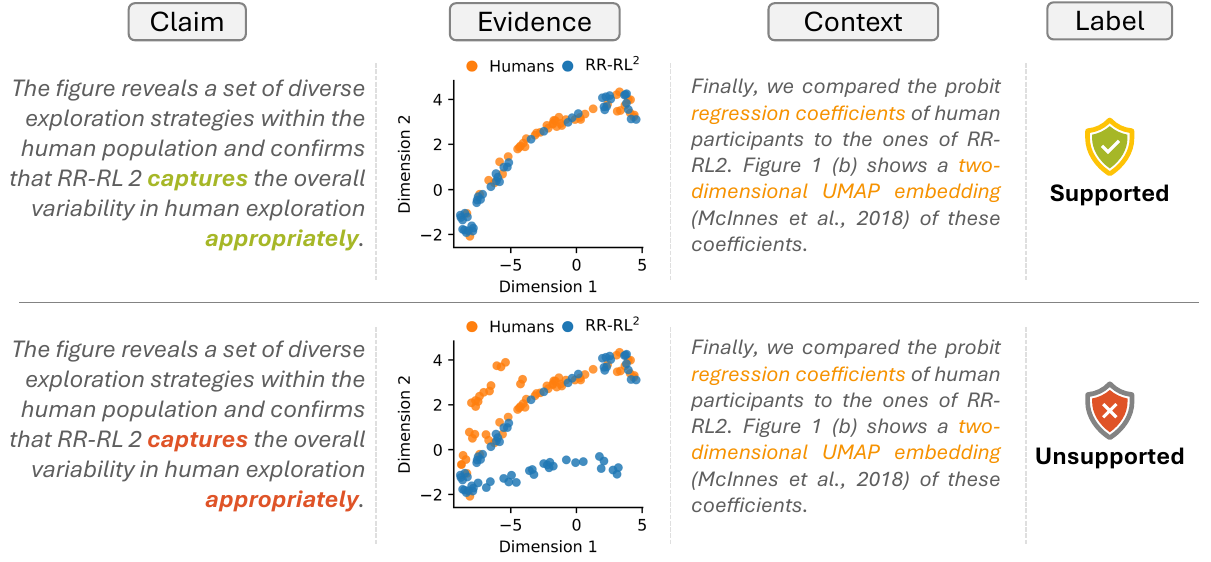}
    \caption{
    An example from our dataset involves creating an unsupported claim by adding spurious data points to the figure. The context provides the necessary information to understand the plot in the evidence.
    }
    \label{fig_example}
\end{figure*}

\paragraph{Human Annotation.}
Annotation was carried out by 11 graduate students and researchers, in two stages:
claim-evidence verification and evidence modification.

In the first stage, annotators see a candidate claim, its table or figure, and the preceding sentences of the same paragraph as context (empty if the claim is the first sentence).
%
%
They label each pair as Supported, Subjectively Supported (the claim relies on subjective terms such as ``large margin'' that cannot be checked against the evidence), Unsupported, or Skip (the claim is purely descriptive, problematic, or requires knowledge the annotator lacks).
They also indicate whether the claim can be understood on its own, needs the short context, or requires the full paper.

In the second stage, annotators work only on Supported pairs.
They modify the table or figure using predefined operations so that the claim is no longer supported by the modified evidence.
For tables, they change cell values, swap row or column names, or add and remove rows or columns. 
For figures, they flip graphs, swap legends or category labels, or exchange subgraphs from the same paper.
Annotators may also propose other edits, such as adding spurious data points (Figure~\ref{fig_example}).
For a few claims, no meaningful modification was possible, and only the supported pair was kept.
%
%
Further details are given in~\citet{ho-etal-2026-sciclaimeval}.

\subsection{Dataset Statistics}

\begin{table}[h!]
    \centering
    \caption{Dataset statistics of SciClaimEval.}
    \begin{tabular}{l r r r}
\toprule
\textbf{Property} & \textbf{Val} & \textbf{Test} & \textbf{All Data} \\
\midrule
\multicolumn{4}{l}{\textit{Labels}} \\ 
\#Supported & 389 & 474 & 863 \\
\#Refuted & 338 & 432 & 770 \\

\midrule
\multicolumn{4}{l}{\textit{Modality Type}} \\ 
Table & 469 & 518 & 987 \\
Figure & 258 & 388 & 646 \\

\midrule
\multicolumn{4}{l}{\textit{Context Use}} \\ 
No & 477 & 611 & 1,088\\
Short Context & 142 & 188 & 330 \\
Full Paper & 108 & 107 & 215 \\

\midrule
\multicolumn{4}{l}{\textit{Domain}} \\ 
NLP & 379 & 384 & 763 \\
ML & 157 & 236 & 393 \\
Medicine & 191 & 286 & 477 \\

\midrule
Total examples & 727 & 906 & 1,633 \\
\bottomrule
\end{tabular}

    \label{tab_dataset_stats}
\end{table}

Table~\ref{tab_dataset_stats} reports statistics for the validation set, the test set, and the full dataset, broken down by modality (table or figure), context use, and domain. 
The split into validation and test sets is balanced with respect to context use and evidence modification operations.
%
The dataset contains \numedit{863} supported and \numedit{770} refuted samples. 
Although each supported claim is meant to have a refuted counterpart, \numedit{93} have none, because no modification of the evidence could be found that was both logical and challenging.

Note that the statistics in Table~\ref{tab_dataset_stats} differ slightly from those reported in~\citet{ho-etal-2026-sciclaimeval}. Following feedback from participants and further analysis, we removed a small number of samples to improve the quality of the dataset.
%
Both the original version (v1.0) and the updated version (v1.1) are available on our Hugging Face page\footnote{https://huggingface.co/datasets/alabnii/sciclaimeval-shared-task} for reproducibility.

\subsection{Dataset Analysis}

\begin{figure}
    \centering
    \includegraphics[width=0.95\linewidth]{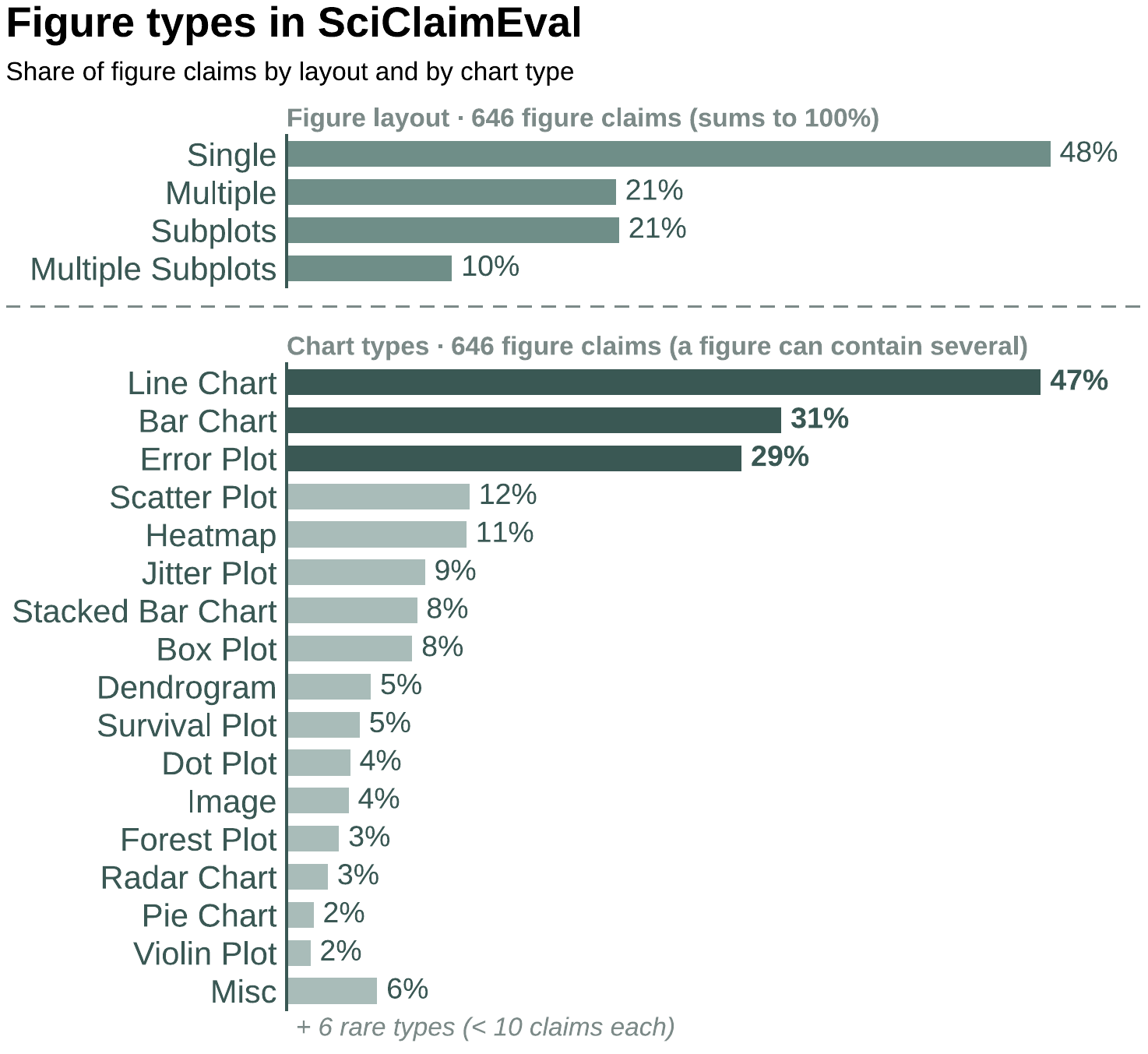}
    \caption{Distribution of all figure types in test and dev split.}
    \label{fig:figuretypes}
\end{figure}

In addition to the analyses of Text Lengths, Evidence-Modifying Operations, and Pixel Changes in Figure Evidence Modification presented in~\citet{ho-etal-2026-sciclaimeval}, we extend the analysis to include chart types.
We manually categorized all figures in the development and test sets. First, we classified figures into four categories based on their plot structure: Single (a single plot), Multiple (multiple unlabeled plots), Subplots (multiple labeled plots, e.g., (a), (b)), and Multiple Subplots (a combination of multiple plots and subplots).
Each individual plot, including those within multiple plots or subplots, may contain a combination of different chart types. The most common combinations are bar charts with error bars or margins. For example, such a figure is categorized as single in terms of plot structure but assigned both bar chart and error plot as chart types.
%

Figure~\ref{fig:figuretypes} shows the distribution of plot structures and chart types across the entire SciClaimEval dataset (development and test splits). We observe a clear bias toward single plots, with line charts, bar charts, and error plots being the most prevalent chart types. Error plots rank among the top three primarily because they frequently appear alongside line and bar charts, but rarely occur independently.


\subsection{Task Design}

SciClaimEval comprises two subtasks.

\paragraph{Subtask 1: Claim-Label Prediction.} 
This is the main subtask.
Given a claim, a piece of evidence (a table or a figure) with its caption, and the contextual information, systems predict whether the claim is supported or refuted by the evidence.
This subtask evaluates whether systems can understand a claim, interpret the evidence, and judge whether the two are consistent.


\paragraph{Subtask 2: Claim-Evidence Prediction.} 
Given a claim, its context, and the two versions of its evidence (the original and the modified one), systems select the version that supports the claim.
The two versions often differ by a single cell or legend entry, so this subtask tests whether systems can find the detail on which the claim depends.

\subsection{Evaluation Metrics}

\paragraph{Subtask 1.}
We report precision, recall, macro-F1, accuracy, and pair accuracy, and use the last two as primary metrics.
Pair accuracy counts a claim as correct only if both of its samples, supported and refuted, are classified correctly.
%
Claims without a refuted counterpart are excluded from this metric.

\paragraph{Subtask 2.}
We report accuracy, the proportion of claims for which the supporting evidence is correctly selected.
The random baseline is 50\%.

\section{Participating Systems}

Fourteen teams submitted 177 runs to SciClaimEval, with thirteen teams taking part in each subtask (Table~\ref{tab:participants}).
This section summarizes the main approaches for each subtask, based on the teams’ system description papers.

\begin{table}[ht!]
    \centering
    \caption{Participating teams, the subtasks they submitted runs to, and their number of runs.}
    \label{tab:participants}
    \resizebox{\columnwidth}{!}{%
    \begin{tabular}{lllcr}
    \toprule
    \textbf{Team ID} & \textbf{Group name} & \textbf{Country} & \textbf{Subtasks} & \textbf{\# Runs} \\
    \midrule
    SIB      & SIB Text Mining             & Switzerland    & 1, 2 &  8 \\
    bit-ias9 & Bonn-Juelich Informatics    & Germany        & 1, 2 & 44 \\
    tabr25   & Black Socks                 & Australia      & 1, 2 & 12 \\
    ECS-UoS  & University of Southampton   & UK & 1, 2 &  2 \\
    ditlab   & Denso IT Laboratory         & Japan          & 1, 2 &  5 \\
    trace29  & SEBIS TUM team              & Germany        & 1, 2 &  2 \\
    tlab     & TakasuLab                   & Japan          & 1, 2 &  7 \\
    procrast & Pro\textsuperscript{2}Crast & Japan          & 1, 2 & 27 \\
    SciNex   & SciNex                      & Japan          & 1, 2 &  8 \\
    ProvideQ & ProvideQ                    & Germany        & 1, 2 & 30 \\
    SCIAI19  & VerifyAI                    & Germany        & 1, 2 &  4 \\
    TUMNLP26 & LMU \& TUM NLP team         & Germany        & 2    &  7 \\
    SciTrue  & SciTrue                     & New Zealand    & 1, 2 & 18 \\
    CAISA26  & CAISALab                    & Germany        & 1    &  3 \\
    \midrule
    \multicolumn{4}{l}{\textbf{Total}} & \textbf{177} \\
    \bottomrule
  \end{tabular}
    }
\end{table}
 
\subsection{Subtask 1: Claim-Label Prediction}

Approaches to Subtask 1 followed three main directions.

\paragraph{Exploiting the Paired Structure.}
The most prominent strategy coordinated predictions across the two samples of each claim.
Black Socks, Southampton, and ditlab compared the two evidence versions associated with the same claim and assigned them complementary labels. 
SciTrue combined model support scores with a pairing constraint, while Bonn-Juelich mapped its Subtask 2 selections back to Subtask 1 labels.
By drawing on information beyond a single claim-evidence pair, these approaches achieved the highest scores.
%
%
Their gains, however, do not transfer to real-world verification, where a claim is assessed without a second, contrasting version of its evidence.

\paragraph{Separating Extraction from Judgment.}
Several teams first turned the evidence into text and then judged the claim against it.
%
VerifyAI transcribed images into text for a separate evaluator. 
ProvideQ converted tables into HTML and extracted structured descriptions of figure elements before verification. 
SciNex explored structured prompting and a two-stage variant that supplemented figure images with descriptions generated without seeing the claim. 

\paragraph{Strengthening Verification.}
Other teams made the verification step itself more rigorous.
Bonn-Juelich combined claim decomposition, context retrieval, and iterative checks.
Takasu Lab checked verifiable conditions with Python code, and CAISALab explored executable Prolog queries. 
Pro²Crast re-verified low-confidence predictions against the full paper, and SIB Text Mining used contradiction-focused prompting and fine-tuning on synthetic data.

\subsection{Subtask 2: Claim-Evidence Prediction}

Approaches to Subtask 2 also followed three main directions.

\paragraph{Direct Comparison.}
Most teams presented both candidates to the model at once.
%
%
SciTrue prompted individual models to compare the candidates directly, or fused their scores in an ensemble.
Bonn-Juelich built a more elaborate joint pipeline, combining claim decomposition, discrepancy reports, iterative verification, and candidate-order checks, with selective human review. 
Pro²Crast supplemented an initial comparison with opposing arguments or interactive debate before a judge selected the evidence, while ditlab explored direct selection across evidence formats and context settings.

\paragraph{Independent Verification, then Selection.}
Other teams first assessed each candidate on its own.
Takasu Lab generated separate verification traces, including deterministic checks and intermediate reasoning, which a text-based selector then compared. 
Southampton and SIB Text Mining used conditional combinations: they initially verified candidates independently and escalated ambiguous cases to joint comparison. 
This allowed them to retain independent judgments when these clearly distinguished the candidates and reserve additional comparison for uncertain decisions.

\paragraph{Separating Extraction from Judgment.}
As in Subtask 1, several teams first extracted information from the evidence, but used it to support a joint decision rather than to label each candidate.
%
VerifyAI compared independently generated transcripts, while SciNex jointly evaluated claim-blind descriptions alongside the original images. 
LMU \& TUM NLP explored selection from structured figure descriptions, and ProvideQ separately grounded each figure before comparing the extracted information. These approaches used separate extraction to support a joint decision, rather than independently assigning support labels to the candidates.

\section{Experimental Results}

Fourteen teams submitted 177 runs: 106 for Subtask 1 and 71 for Subtask 2, with thirteen teams participating in each subtask.
The official ranking is based on runs using the PNG version of the evidence, since structured formats such as JSON, LaTeX, and HTML are likely easier for language models to interpret. Participants could also submit optional runs using these structured formats.
Note that all scores reported here are evaluated on the test set of our dataset.

\subsection{Subtask 1: Claim-Label Prediction}

Table~\ref{tab:subtask1_png} shows the best-performing runs\footnote{The results of other runs are available on our task website: \url{https://sciclaimeval.github.io/}.} of each team for Subtask 1, where systems are ranked by pair accuracy. Bonn-Juelich Informatics and SciTrue achieved the highest pair accuracy of 98.1\%, with SciTrue slightly ahead on all other metrics. Both systems clearly outperformed the other participants, with the third-ranked team trailing by 8.1 percentage points in pair accuracy.

\begin{table}[ht]
\centering
\small
\caption{Best run of each team on Subtask 1 (PNG evidence), ranked by pair accuracy. $^\dagger$ indicates runs that use the paired structure of the test data.}
\label{tab:subtask1_png}
\resizebox{\columnwidth}{!}{%
\begin{tabular}{lccccc}
\toprule
Team & P & R & Macro-F1 & Acc. & Pair Acc. \\
\midrule
SciTrue$^\dagger$ & \textbf{98.0} & \textbf{98.0} & \textbf{98.0} & \textbf{98.0} & \textbf{98.1} \\
Bonn-Juelich Informatics$^\dagger$ & 97.2 & 97.3 & 97.2 & 97.2 & \textbf{98.1} \\
Black Socks$^\dagger$ & 94.1 & 93.7 & 93.8 & 93.8 & 90.0 \\
Pro²Crast$^\dagger$ & 87.2 & 87.2 & 87.2 & 87.2 & 86.8 \\
University of Southampton$^\dagger$ & 85.8 & 85.9 & 85.8 & 85.9 & 85.9 \\
Denso IT Laboratory$^\dagger$ & 74.1 & 74.1 & 74.1 & 74.1 & 75.9 \\
TakasuLab & 86.7 & 86.4 & 86.5 & 86.5 & 73.8 \\
SEBIS TUM & 72.7 & 72.7 & 72.7 & 72.7 & 72.9 \\
VerifyAI & 85.3 & 85.4 & 85.3 & 85.3 & 72.2 \\
ProvideQ$^\dagger$ & 84.8 & 84.9 & 84.9 & 84.9 & 71.1 \\
SIB Text Mining & 81.3 & 81.3 & 81.1 & 81.1 & 64.1 \\
\textit{Baselines} & 80.4 & 79.5 & 79.8 & 79.4 & 61.1 \\
SciNex & 77.5 & 77.5 & 77.4 & 77.4 & 57.4 \\
CAISALab & 62.3 & 59.8 & 58.2 & 60.8 & 26.9 \\
\bottomrule
\end{tabular}
}
\end{table}

For Subtask 1, the task design did not explicitly prohibit systems from exploiting the paired structure of the test data. Based on our review of the system descriptions, the six top-ranked systems used this information, typically by assigning complementary labels to the two evidence versions of each claim. Therefore, their results reflect a setting where both evidence versions are available, rather than the intended setting where each claim-evidence pair is evaluated independently. 
Among systems that did not exploit paired information, TakasuLab achieved the highest pair accuracy of 73.8\%, which is 24.3 percentage points lower than the top-performing systems. A similar performance drop was observed by the two leading teams on the development set when paired information was removed, with pair accuracy decreasing from 95.7\% to 74.4\% and from 94.3\% to 82.7\%, respectively. Additionally, Bonn-Juelich Informatics incorporated selective human review, meaning its runs were not fully automatic. In Table~\ref{tab:subtask1_png}, runs that exploited paired information are marked with $\dagger$.


Tables~\ref{tab:subtask1_json} and~\ref{tab:subtask1_tex} show the results for Subtask 1 using the optional structured formats, JSON and \LaTeX/HTML, respectively. Surprisingly, runs using these formats did not clearly outperform those using PNG evidence. One possible reason is that most teams focused on multimodal models, in line with the primary goal of the task, and optimized their systems for images. Additionally, fewer runs were submitted using these formats, limiting direct comparisons. However, other factors may also contribute to these results, and no firm conclusions can be drawn.

\begin{table}[ht]
\centering
\small
\caption{Subtask 1 results (JSON evidence).}
\label{tab:subtask1_json}
\begin{tabular}{lccccc}
\toprule
Team & P & R & Macro-F1 & Acc. & Pair Acc. \\
\midrule
SciTrue & \textbf{98.1} & \textbf{98.1} & \textbf{98.1} & \textbf{98.1} & \textbf{98.4} \\
Black Socks & 93.1 & 93.2 & 93.2 & 93.2 & 88.7 \\
Pro²Crast & 83.6 & 83.7 & 83.6 & 83.7 & 83.6 \\
TakasuLab & 85.0 & 84.9 & 84.9 & 85.0 & 71.3 \\
SciNex & 78.3 & 78.1 & 77.8 & 77.8 & 59.7 \\
CAISALab & 65.7 & 64.7 & 64.4 & 65.2 & 32.6 \\
\bottomrule
\end{tabular}
\end{table}

\begin{table}[ht]
\centering
\small
\caption{Subtask 1 results (TeX/HTML evidence).}
\label{tab:subtask1_tex}
\begin{tabular}{lccccc}
\toprule
Team & P & R & Macro-F1 & Acc. & Pair Acc. \\
\midrule
SciTrue & \textbf{98.1} & \textbf{98.1} & \textbf{98.1} & \textbf{98.1} & \textbf{98.4} \\
Bonn-Juelich Informatics & 96.7 & 96.7 & 96.7 & 96.7 & 97.7 \\
ProvideQ & 89.9 & 90.0 & 90.0 & 90.0 & 82.4 \\
\bottomrule
\end{tabular}
\end{table}

\subsection{Subtask 2: Claim-Evidence Prediction}

Table~\ref{tab:subtask2_png} presents the best-performing runs of each team for Subtask 2, where systems are ranked by accuracy. SciTrue achieved the highest accuracy, outperforming Bonn-Juelich Informatics by a narrow margin of 0.3 percentage points. The gap between the top-performing systems and the other participants was smaller than in Subtask 1, with the third-ranked team, the University of Southampton, achieving 96.3\%, 2.1 percentage points behind SciTrue.

\begin{table}[ht]
\centering
\small
\caption{Best run of each team on Subtask 2 results (PNG evidence).}
\label{tab:subtask2_png}
\begin{tabular}{lc}
\toprule
Team & Accuracy \\
\midrule
SciTrue & \textbf{98.4} \\
Bonn-Juelich Informatics & 98.1 \\
University of Southampton & 96.3 \\
TakasuLab & 94.0 \\
ProvideQ & 91.7 \\
SIB Text Mining & 90.5 \\
SciNex & 88.2 \\
\textit{Baselines} & 81.9 \\
LMU \& TUM NLP & 81.5 \\
Pro²Crast & 80.8 \\
Denso IT Laboratory & 61.3 \\
SEBIS TUM & 60.6 \\
\bottomrule
\end{tabular}
\end{table}

The baseline achieves 81.9\% accuracy on Subtask 2, compared to 61.1\% pair accuracy on Subtask 1, with the six top-performing systems all achieving scores above 90\%. This performance difference is expected, as selecting the better-supported evidence from two nearly identical pieces of evidence requires only a relative judgment.


Tables~\ref{tab:subtask2_json} and~\ref{tab:subtask2_tex} present the results for Subtask 2 using JSON and \LaTeX/HTML formats, respectively. Similar to Subtask 1, runs using structured formats did not show a clear advantage over those using PNG evidence. This may be because most participating teams prioritized multimodal approaches and optimized their systems for image-based evidence. The smaller number of submissions using structured formats also limits the comparison. Therefore, further investigation is needed to determine whether structured representations provide advantages over image-based evidence for this task.


\begin{table}[ht]
\centering
\small
\caption{Subtask 2 results (JSON evidence).}
\label{tab:subtask2_json}
\begin{tabular}{lc}
\toprule
Team & Accuracy \\
\midrule
TakasuLab & \textbf{93.5} \\
SciNex & 89.3 \\
Black Socks & 84.5 \\
LMU \& TUM NLP & 83.6 \\
VerifyAI & 66.7 \\
Pro²Crast & 66.7 \\
Denso IT Laboratory & 65.5 \\
\bottomrule
\end{tabular}
\end{table}

\begin{table}[ht]
\centering
\small
\caption{Subtask 2 results (TeX/HTML evidence).}
\label{tab:subtask2_tex}
\begin{tabular}{lc}
\toprule
Team & Accuracy \\
\midrule
Bonn-Juelich Informatics & \textbf{97.7} \\
\bottomrule
\end{tabular}
\end{table}

\section{Analyses}

\subsection{Accuracy Across Evidence Modifications}

\begin{figure*}
    \centering
    \includegraphics[width=0.85\linewidth]{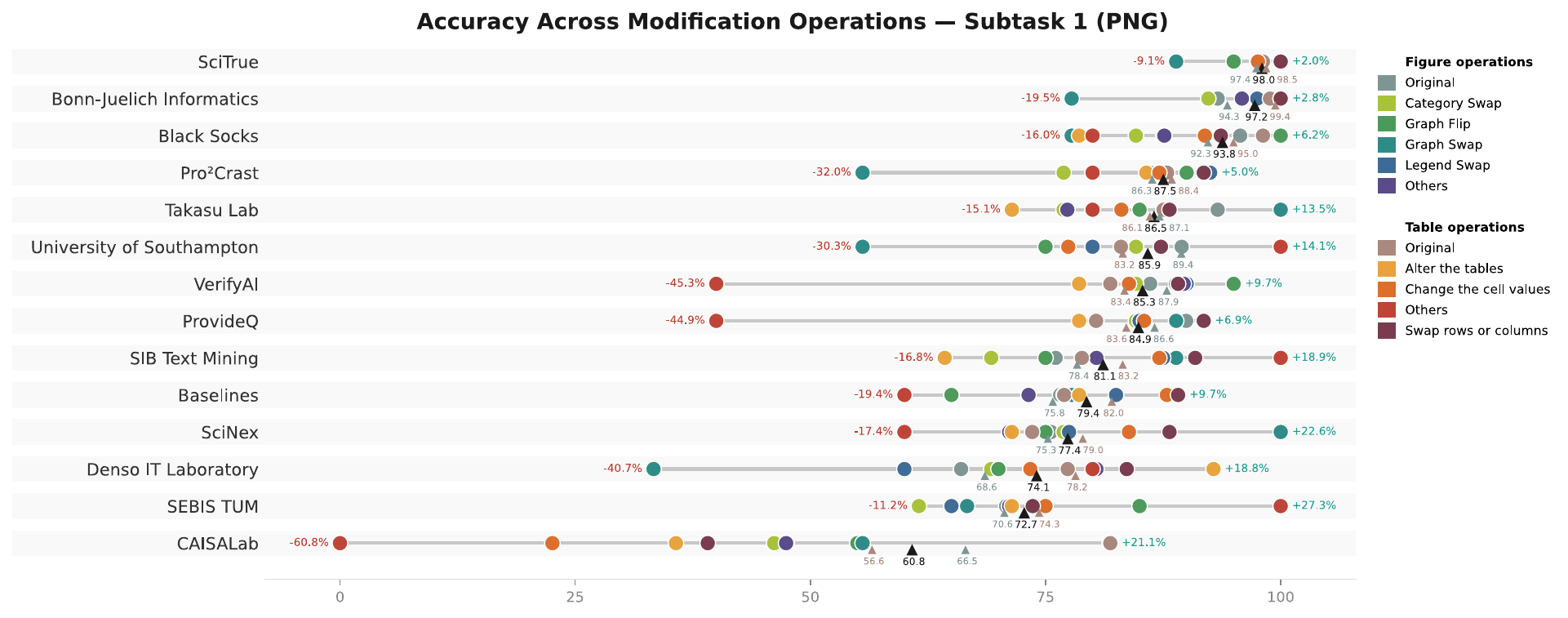}
    \caption{Accuracy on Subtask 1 across evidence modification categories. The red and green percentages indicate performance drops and gains, respectively, compared to the overall average (black triangle marker). The brown and turquoise triangle markers indicate the average accuracy for tables and figures, respectively. 
    }
    \label{fig:subtask1_spread}
\end{figure*}

\begin{figure*}
    \centering
    \includegraphics[width=0.85\linewidth]{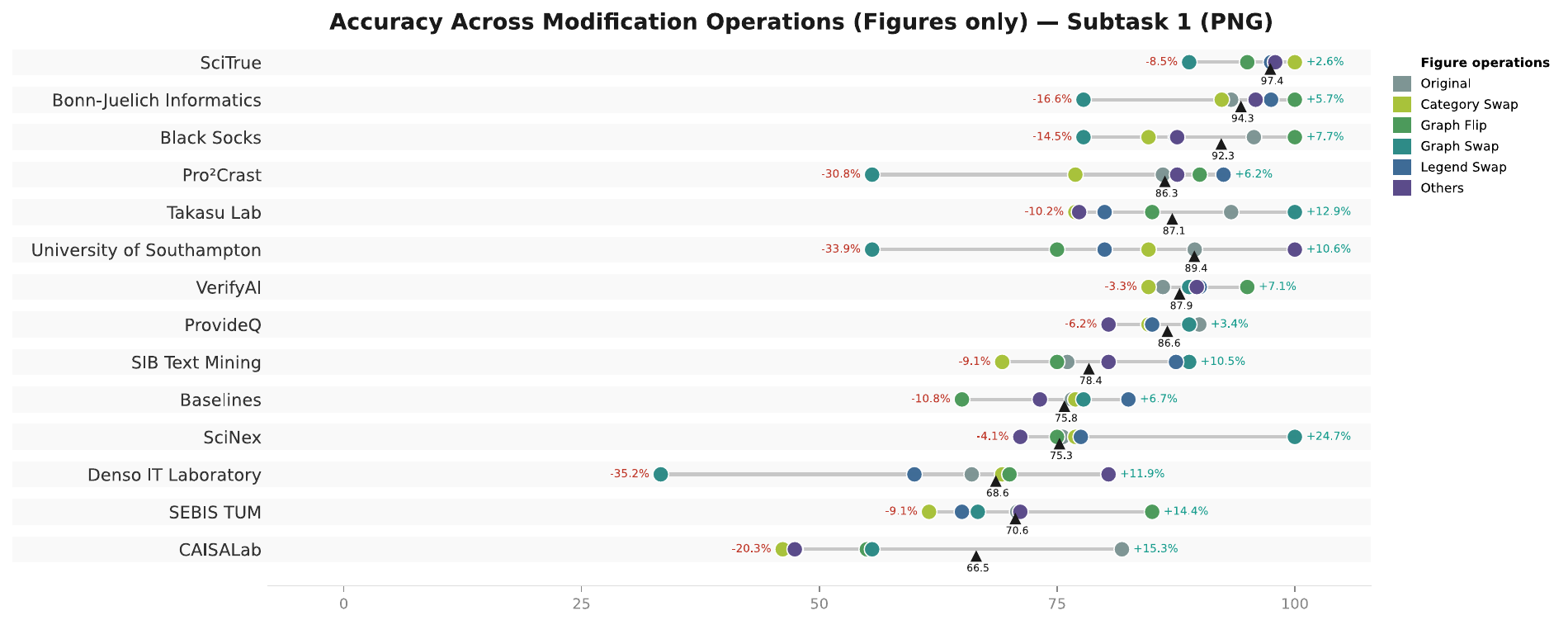}
    \caption{Accuracy on Subtask 1 across figure modification categories. The red and green percentages indicate performance drops and gains, respectively, compared to the average accuracy across all figures (black triangle marker).
    }
    \label{fig:subtask1_spread_figures}
\end{figure*}

\begin{figure*}
    \centering
    \includegraphics[width=0.85\linewidth]{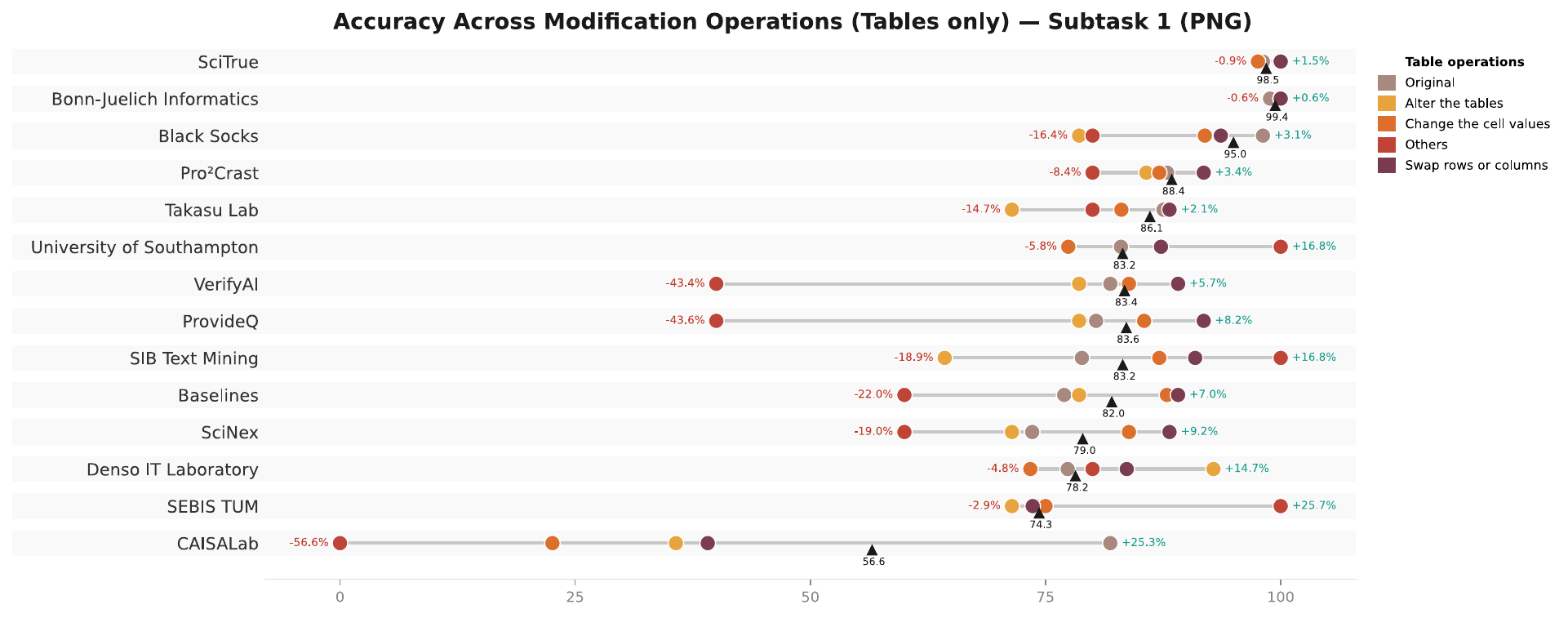}
    \caption{Accuracy on Subtask 1 across table modification categories. The red and green percentages indicate performance drops and gains, respectively, compared to the average accuracy across all tables (black triangle marker).
    }
    \label{fig:subtask1_spread_tables}
\end{figure*}

Figure~\ref{fig:subtask1_spread} shows the accuracy across all modifications applied to tables and figures. We would expect certain modifications to be more difficult to detect than others. For example, swapping rows or columns is likely easier to identify than modifying individual cell values. Interestingly, however, teams struggled with different types of modifications and evidence. While some teams performed better on tables, as expected given their more structured nature compared to figures, others showed a performance drop on tables.

Figure~\ref{fig:subtask1_spread} also reveals substantial variation in performance across modification types, with no single modification consistently challenging all teams. For example, the table modification \textit{others} caused VerifyAI and ProvideQ to perform 45\% below their respective averages, and CAISALab nearly 60\% below its average. In contrast, SciTrue, Bonn-Juelich Informatics, University of Southampton, SIB Text Mining, and SEBIS TUM all achieved 100\% accuracy on the same modification. Similarly, \textit{graph swaps} particularly challenged the top-performing teams, while Takasu Lab and SciNex achieved 100\% accuracy. Notably, performance on \textit{graph swaps} and \textit{category swaps} likely contributed to SciTrue outperforming Bonn-Juelich Informatics in the overall ranking.

Figures~\ref{fig:subtask1_spread_figures} and~\ref{fig:subtask1_spread_tables} present the same analysis as Figure~\ref{fig:subtask1_spread}, but separately for figures and tables, respectively. The results reveal that the top two teams, SciTrue and Bonn-Juelich Informatics, outperformed other teams primarily on tables. While most teams struggled with tables and exhibited large performance variations across modification types, the top two teams remained relatively unaffected, with performance spreads of only around 2\%. Notably, Bonn-Juelich Informatics made only three mistakes across all tables.

For figures, ProvideQ and VerifyAI achieved the smallest performance spreads of 9.6\% and 10.4\%, respectively, followed by SciTrue at 11.1\%. This difference likely played a key role in determining the winner of this year's SciClaimEval task, as Bonn-Juelich Informatics exhibited a much larger spread of 22.3\%, including a performance drop of 16.6\% on \textit{graph swaps} alone.

These large performance variations further highlight that no team developed a solution that performed consistently well across all evidence types. Instead, teams demonstrated strong performance on specific types of claim–evidence pairs.

\begin{figure*}
    \centering
    \includegraphics[width=0.85\linewidth]{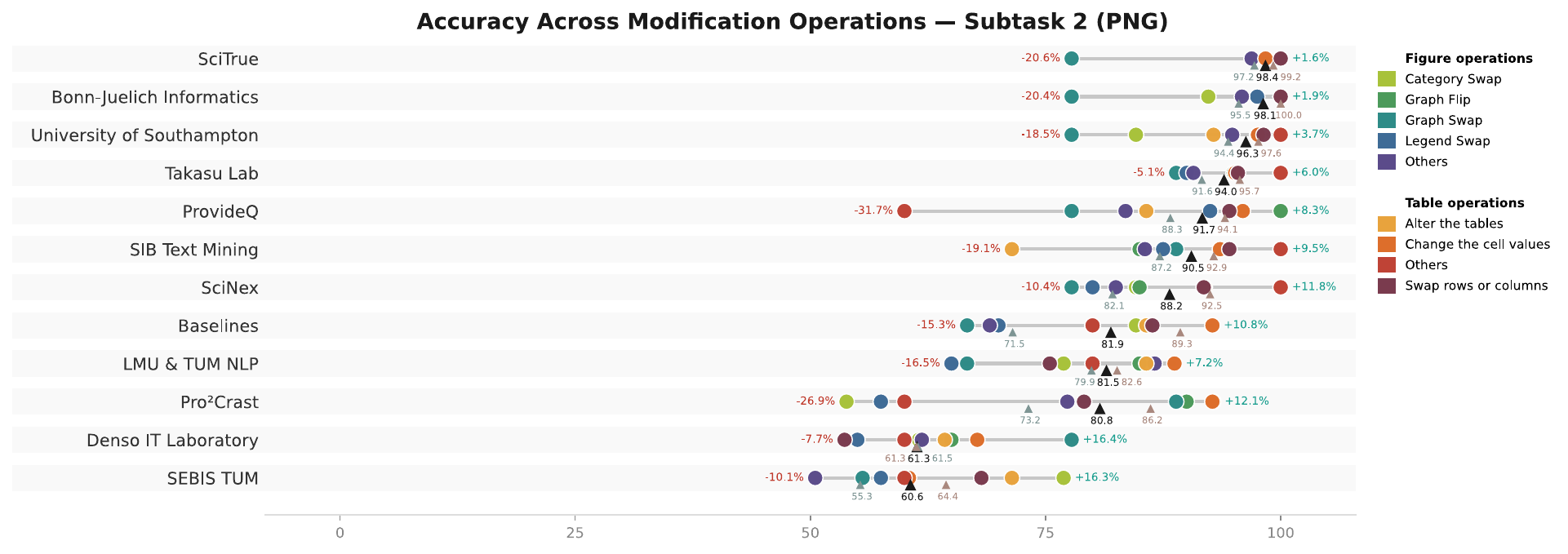}
    \caption{Accuracy on Subtask 2 across evidence modification categories. The red and green percentages indicate performance drops and gains, respectively, compared to the overall average accuracy (black triangle marker). The brown and turquoise triangle markers indicate the average accuracy for tables and figures, respectively.
    }
    \label{fig:subtask2_spread}
\end{figure*}

Figure~\ref{fig:subtask2_spread} presents the same Cleveland plot as Figure~\ref{fig:subtask1_spread}, but for Subtask 2 submissions. While the results largely align with those of Subtask 1, some interesting outliers emerge. For example, SciTrue, Bonn-Juelich, and Southampton achieved identical scores on \textit{graph swap} modifications. In contrast, on Subtask 1, SciTrue performed significantly better, while Southampton performed significantly worse on this modification type.


\subsection{Accuracy Across Figure Types}

\begin{figure*}
    \centering
    \includegraphics[width=0.85\linewidth]{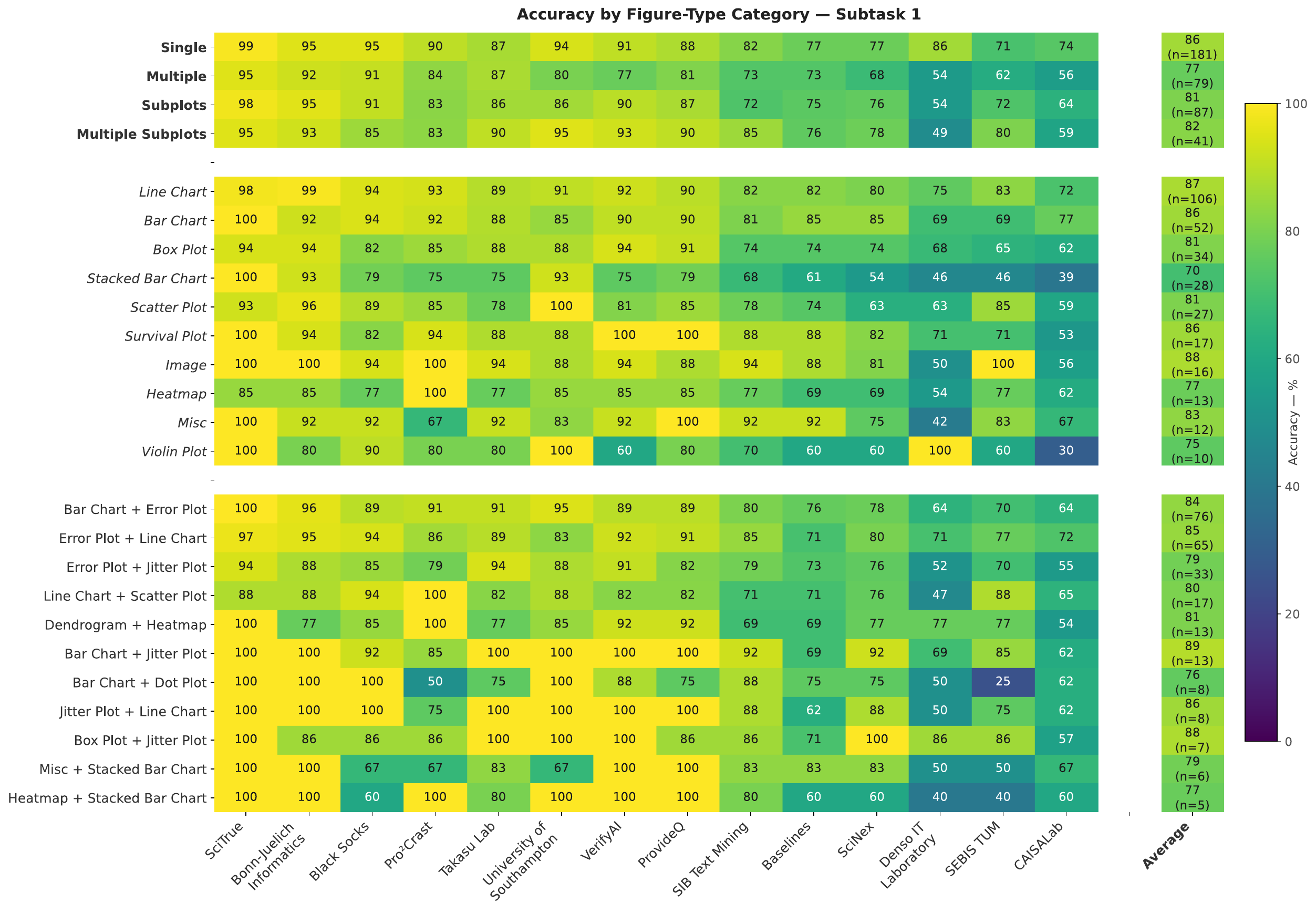}
    \caption{Best-run accuracy on Subtask 1 across figure types. 
    \textit{Single} refers to a single plot, which may combine multiple chart types (e.g., a bar chart with error bars). \textit{Multiple} refers to multiple unlabeled plots, \textit{Subplots} refers to multiple labeled plots (e.g., (a), (b)), and \textit{Multiple Subplots} refers to a combination of labeled and unlabeled plots.
    Italicized categories (middle part) represent chart types across all figure structures (Single, Multiple, Subplots, and Multiple Subplots). The bottom category, \textit{Combination}, refers to figures containing two or more chart types, regardless of their structure.
    }
    \label{fig:performance_per_figuretypes}
\end{figure*}

Figure~\ref{fig:performance_per_figuretypes} shows the accuracy of each team's best run across figure categories. Note that the categorization was performed independently of the claims. In other words, low performance on a particular chart type, such as a violin plot, does not necessarily indicate that an approach struggles to interpret that type. Instead, it indicates lower performance when that chart type is present in the figure, regardless of whether the claim refers to it. Furthermore, unlike in Figure~\ref{fig:figuretypes}, each chart type was counted only once per figure for evaluation.

Figure~\ref{fig:performance_per_figuretypes} reveals a clear and consistent performance gap between the bottom six runs and the top-performing teams. However, no clear correlation between chart types and performance emerges. Stacked bar charts generally lead to lower performance but pose no difficulty for SciTrue. Similarly, heatmaps prove challenging for most teams except Pro$^2$Crast.


Another observation from Figure~\ref{fig:performance_per_figuretypes} is that visual noise (additional irrelevant information) appears to have little impact, as single figures outperform multiple subplots and subplots by only 4\% and 5\% on average, respectively. Although figures containing multiple plots show a substantial average performance drop of up to 9\%, this is primarily driven by lower-performing teams.
These findings suggest that no single approach performs equally well across all chart types, further supporting the observations in Figures~\ref{fig:subtask1_spread} and~\ref{fig:subtask2_spread}.


\subsection{Label and Evidence Selection Bias}

\begin{figure}
    \centering
    \includegraphics[width=0.95\linewidth]{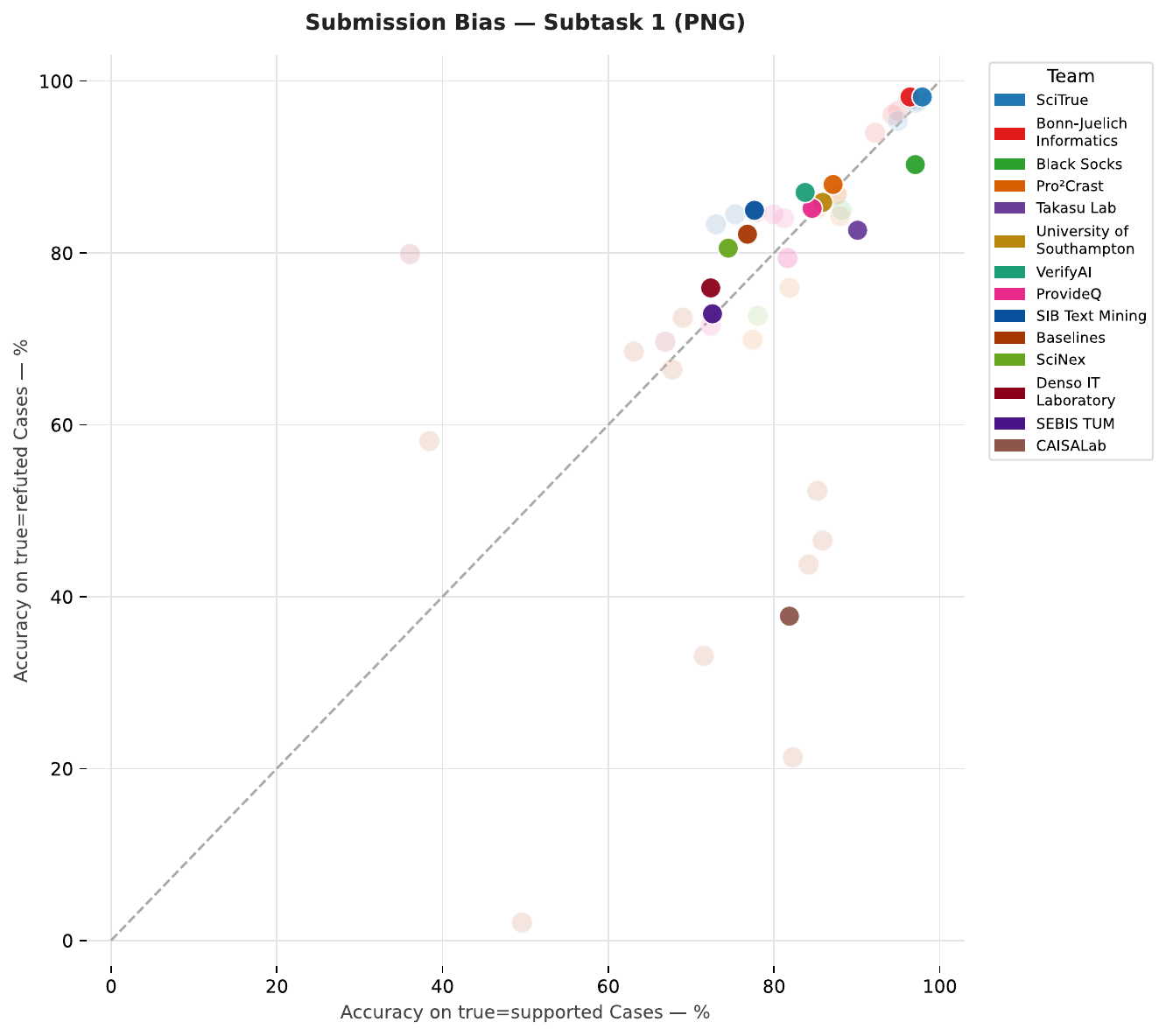}
    \caption{Label prediction bias across all runs in Subtask 1.}
    \label{fig:bias_t1}
\end{figure}

\begin{figure}
    \centering
    \includegraphics[width=0.95\linewidth]{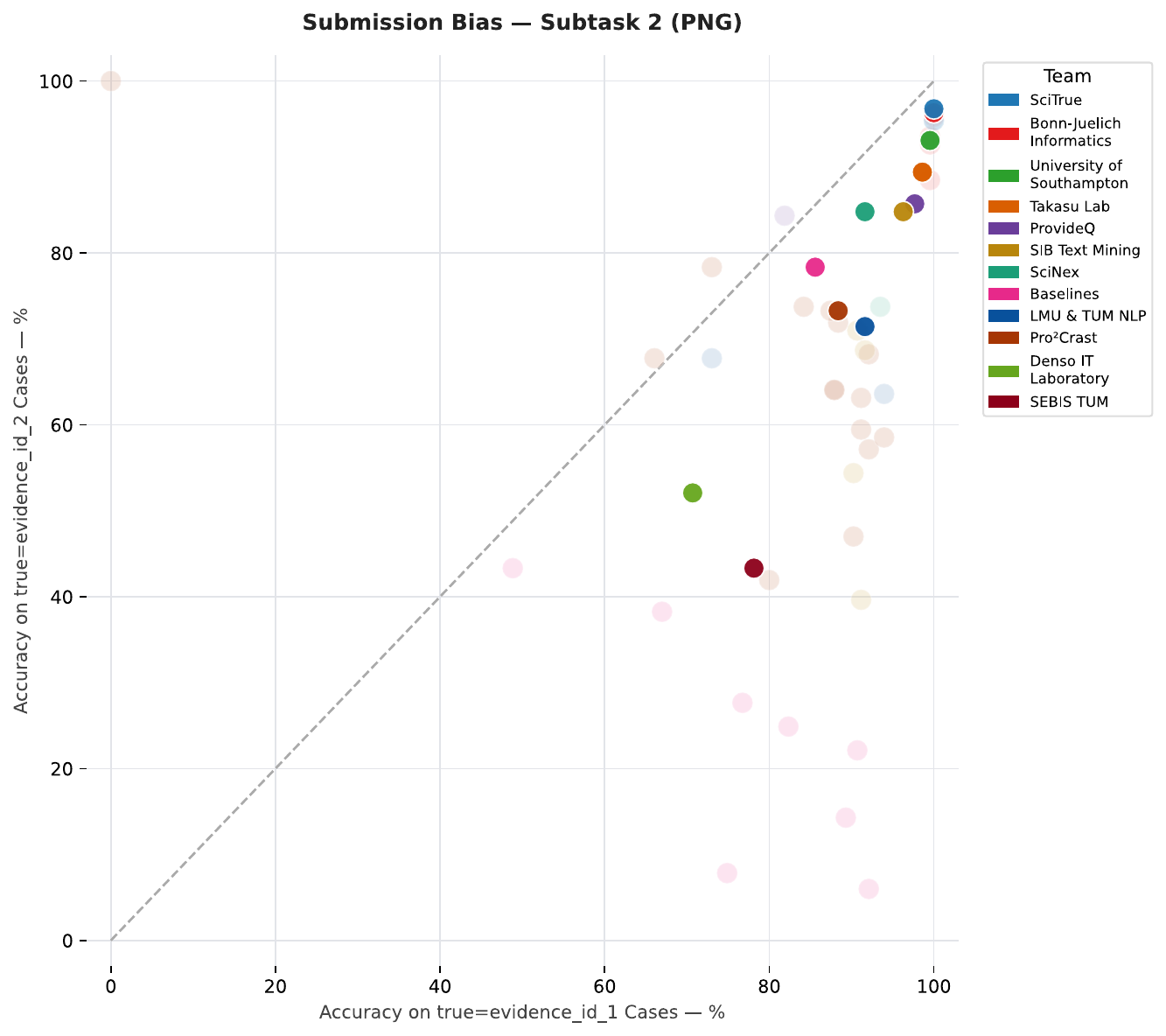}
    \caption{Evidence selection bias across all runs in Subtask 2.}
    \label{fig:bias_t2}
\end{figure}

\begin{figure*}
    \centering
    \includegraphics[width=0.99\linewidth]{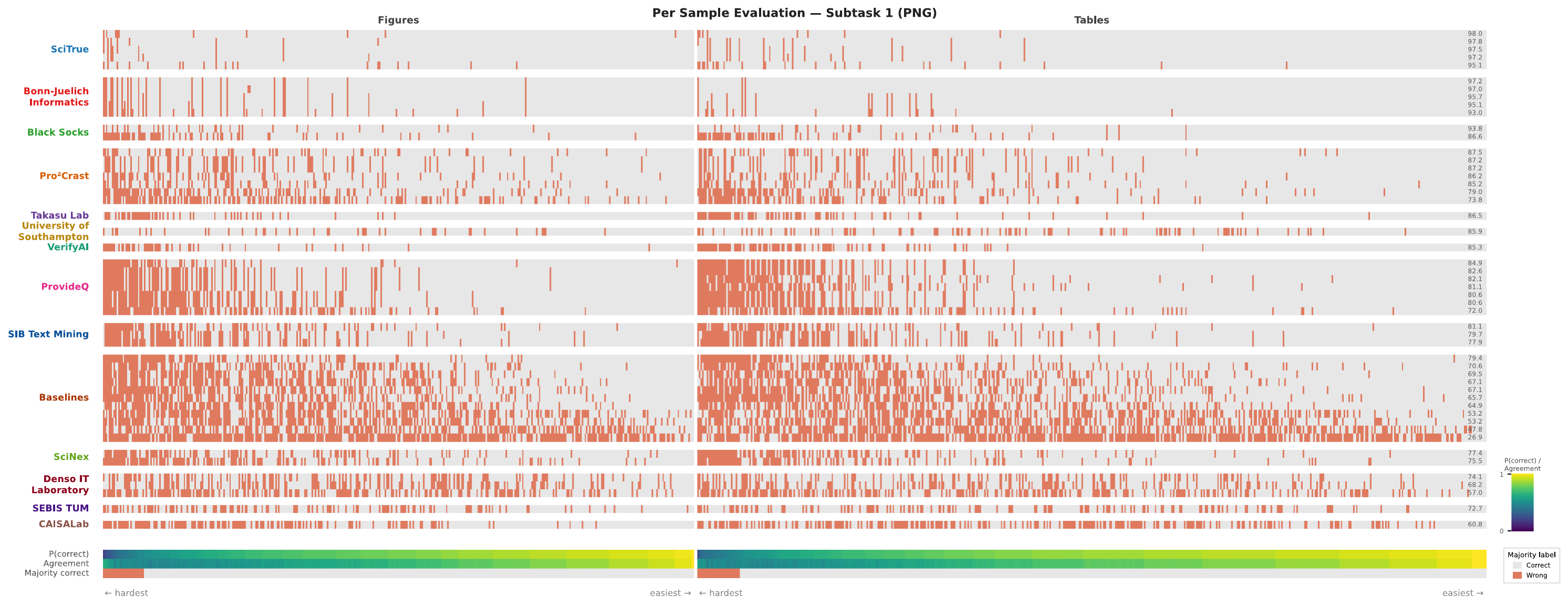}
    \caption{Full evaluation results for each submitted run and claim–evidence sample. Red and gray markers indicate incorrect and correct predictions, respectively. Queries are sorted by consensus from hardest (left, most runs predicted incorrectly) to easiest (right, most runs predicted correctly), with figure samples on the left and table samples on the right. The rightmost gray labels show the average accuracy of each run. 
    At the bottom, $P(\text{correct})$ represents the fraction of runs with correct predictions for each sample, while \textit{Agreement} indicates the agreement among all runs on the predicted label. \textit{Majority correct} is marked red if more than half of the runs predicted incorrectly and gray otherwise.
    }
    \label{fig:subtask1_perquery}
\end{figure*}

\begin{figure*}
    \centering
    \includegraphics[width=0.8\linewidth]{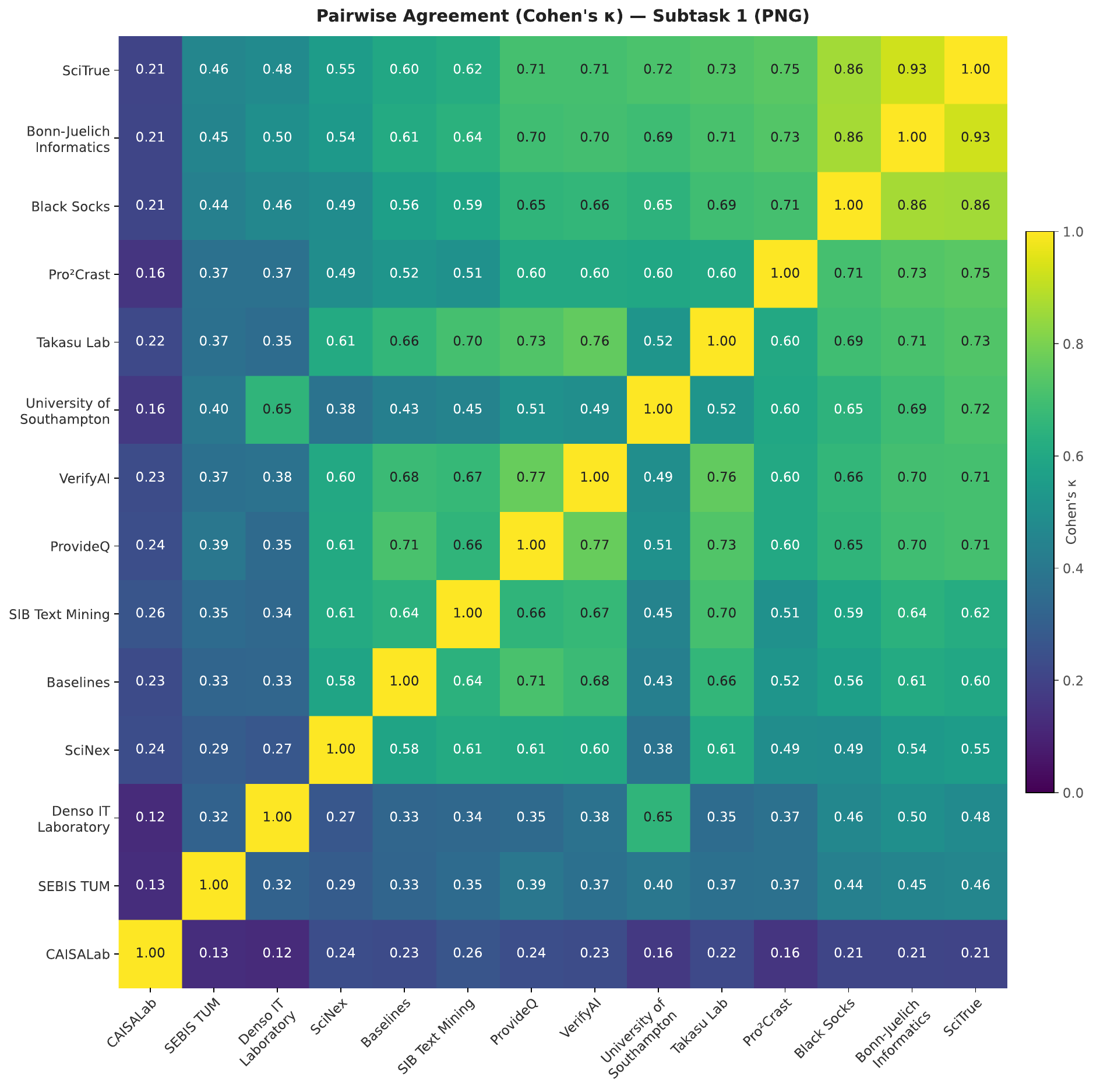}
    \vspace*{-0.5cm}
    \caption{Similarity (Cohen's $\kappa$) between each team's best-performing run, with runs sorted by accuracy.}
    \label{fig:run_similarity_task1}
\end{figure*}

Figures~\ref{fig:bias_t1} and~\ref{fig:bias_t2} show the biases toward selected labels in Subtasks 1 and 2, respectively. In Subtask 1, no clear bias toward either supported or refuted labels emerges across teams and submitted runs. Only a few outliers, particularly several baselines, exhibit a clear bias toward supported cases. CAISALab's best-performing run also shows a substantial bias toward predicting supported labels.

Surprisingly, almost all teams and submitted runs, except for one run by Pro$^2$Crast, favored the first provided evidence in Subtask 2. The reason for this bias remains unclear. One possible explanation is that processing two images may affect the attention mechanisms of MLLMs, causing them to default to the first input. Notably, no team's best-performing run was able to overcome this bias, despite the wide variety of MLLMs and pipeline approaches employed.



\subsection{Sample Difficulty and Run Similarity}

Figure~\ref{fig:subtask1_perquery} shows the evaluation results of all submitted runs across all claim–evidence samples. We observe that most teams struggled with similar samples and performed well on the same easier samples, as indicated by the red regions in the bottom-left and gray regions in the top-right of the matrix. However, only a relatively small proportion of samples were incorrectly labeled by most submitted runs. In fact, no single sample was incorrectly labeled by all runs. This pattern is consistent with what we would expect from a high-quality dataset without strong systematic biases.

One notable outlier is the University of Southampton's run, which does not show an increasing number of incorrect predictions toward the harder samples on the left. In other words, this run performed relatively well on samples that most other teams struggled with. This distinction becomes even more apparent in Figure~\ref{fig:run_similarity_task1}, which presents a pairwise agreement matrix based on Cohen's $\kappa$. The University of Southampton is again a clear outlier, showing its third-highest agreement with Denso IT Laboratory despite outperforming it by 11.8\% in accuracy (and 10\% in pair accuracy). 
Figure~\ref{fig:performance_per_figuretypes} also showed this team outperformed most of the other team's in its vicinity on the two most challenging figure types stacked bar charts and violin plots. 
However, it remains unclear why the University of Southampton's approach exhibits relatively low agreement with other similarly high-performing teams.


\subsection{Impact of Pair Information on Performance}

\begin{figure}
    \centering
    \includegraphics[width=0.95\linewidth]{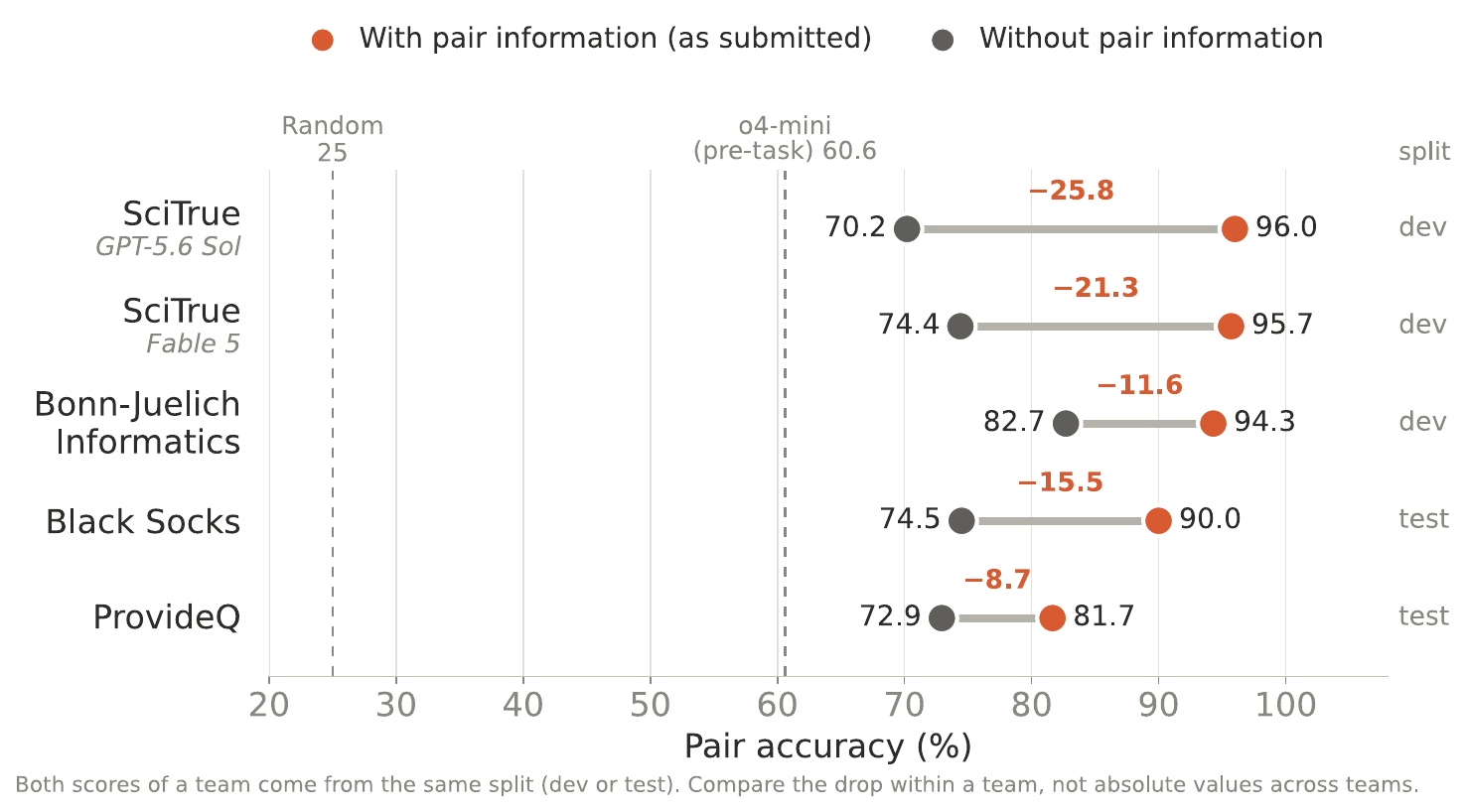}
    \caption{Reported performance gains from exploiting pair information. Not all teams reported these gains, and some reported results only on the development (dev) split. SciTrue reported multiple gains depending on the underlying LLM.
    }
    \label{fig:pair_drop}
\end{figure}

As earlier pointed out, one of the most common themes to tackle subtask 1 was the exploitation of pair information. This is a clear shortcoming of our task design that we overlooked in favour of a more robust evaluation. However, it is immediately obvious that this feature made the task significantly easier. In fact, we designed subtask 2 specifically to illustrate the different challange between subtask 1 and 2.
As humans, analyzing an example, such as in Figure~\ref{fig_example}, is also significantly easier with both evidence figures present as it allows us to determine changes and guide our judgement visually first before understanding the claim.

Most teams acknowledged that exploiting pair information leads to unrealistic evaluations and self-reported results without utilizing pairs at least for some earlier iterations of their run submissions.
Figure~\ref{fig:pair_drop} shows these self-reported performance gains. Using this information made the task substantially easier for all teams and, in extreme cases, effectively reduced Subtask 1 to Subtask 2. However, these findings highlight an important conclusion for SciClaimEval: scientific claim verification remains challenging, and the task itself is far from solved.

For example, SciTrue reported that the propriatory frontier models Claude's Fable 5 achieved only 74.4\% pair accuracy and OpenAI's GPT-5.6 Sol only 70.2\% pair accuracy on the development set. In fact, all teams that exploited the pair structure and reported results without using pair information observed substantial drops in pair accuracy, ranging from 8.7\% (ProvideQ on the test set) to 25.8\% (SciTrue's evaluation with GPT-5.6 Sol on the development set).

These findings suggest that the high performance achieved by some teams was not necessarily due to stronger verification capabilities, but rather to access to pair information that would typically be unavailable in real-world scenarios. Consequently, these results provide a strong motivation for future iterations of SciClaimEval that remove pair information.


\section{Conclusion}
This paper presented an overview of SciClaimEval, a pilot shared task at NTCIR-19 on cross-modal scientific claim verification using evidence from tables and figures. We introduced a new benchmark dataset covering multiple scientific domains, with contrastive evidence pairs, and summarized the participation and system performance across two subtasks. The results demonstrate the strong potential of MLLMs for this task, particularly when combined with carefully designed inference pipelines and effective use of contrastive evidence pairs. 
However, the substantial performance gap between systems using contrastive pairs and those evaluating individual claim–evidence pairs highlights the remaining challenges in practical scientific claim verification. We hope that SciClaimEval will serve as a valuable benchmark for advancing research on reliable multimodal scientific claim verification.

\section{Discussion \& Future Directions}

Overall, we consider SciClaimEval a success. This initial edition attracted significant community interest and the highest number of participants at NTCIR-19. However, it also revealed shortcomings in the dataset design that allowed teams to exploit information, making the task substantially easier than intended. Although the resulting high scores may raise questions about the need for further scientific claim verification tasks, the self-reported performance drops without pair information demonstrate that real-world scientific claim verification remains challenging.

Most claim–evidence mismatches in SciClaimEval are relatively easy for humans to identify, even without paired evidence, and their prevalence in actual scientific writing remains unclear. In practice, more challenging cases arise in three scenarios. First, verification may require domain expertise, such as interpreting equations, medical scans, or crystal structures. Second, evidence may be difficult to access or scattered across lengthy documents. For example, verifying claims against cited works often requires retrieving and examining entire papers. Third, claim–evidence relationships are rarely strictly binary and may require more nuanced judgments than our current task allows.

Future iterations of SciClaimEval will address these limitations. 
First, we will eliminate access to pair information at test time to prevent the shortcuts observed in this edition. 
Second, we will expand the dataset beyond AI, NLP, and medicine to domains such as materials science and physics, incorporating additional evidence types, including equations, medical scans, and crystal structure visualizations. 
Third, we will introduce cited papers as evidence sources, addressing the time-consuming and error-prone process of verifying citations, particularly relevant with the rise of AI-assisted scientific writing. 
Finally, we plan to move beyond binary classification by introducing more nuanced labels or reformulating the task, reducing reliance on shortcuts and enabling a more comprehensive evaluation of scientific reasoning.

\section*{Acknowledgments}
This work was supported by JSPS KAKENHI Grant Number 24K03231,
the French National Research Agency (ANR) through the FABULEUX project (ANR-26-CE23-6589),
and by the Deutsche Forschungsgemeinschaft (DFG, German Research Foundation) -- 554559555.
F. Boudin was also supported by a visiting professor position at the National Institute of Informatics (NII), Japan.




\bibliographystyle{ACM-Reference-Format}
\bibliography{ntcirsample}

\end{document}